\documentclass[journal]{IEEEtran}
\usepackage{cite}
\usepackage[pdftex]{graphicx}
\usepackage{amsmath}
\usepackage{color}
\usepackage{array}
\usepackage[caption=false,font=footnotesize]{subfig}
\usepackage{url}
\usepackage{booktabs}
\usepackage{wasysym}

\begin{document}
\bstctlcite{IEEEexample:BSTcontrol}

\title{Perivascular Spaces Segmentation in Brain MRI Using Optimal 3D Filtering}

\author{Lucia Ballerini*~\IEEEmembership{Member,~IEEE},
Ruggiero~Lovreglio, 
Maria~del~C.~Vald{\'e}s~Hern{\'a}ndez~\IEEEmembership{Member,~IEEE}, 
Joel~Ramirez, 
Bradley~J.~MacIntosh,
Sandra~E.~Black, and
Joanna~M.~Wardlaw
\thanks{This work was supported by SINAPSE, Fondation Leducq, MRC Disconnected Mind (MR/M013111/1), Age UK (DCM Phase 2), Row Fogo Charitable Trust (R35865, R43412). }
\thanks{L.~Ballerini, M.~del~C.~Vald{\'e}s~Hern{\'a}ndez and J.~M.~Wardlaw are with the Department of Neuroimaging Sciences, University of Edinburgh, UK}
\thanks{R.~Lovreglio is with the Department of Civil and Environmental Engineering, University of Auckland, New Zealand}
\thanks{J.~Ramirez, B.~MacIntosh and S.~E.~Black are with the Hurvitz Brain Sciences Program, LC Campbell Cognitive Neurology Research Unit, Heart and Stroke Foundation Canadian Partnership for Stroke Recovery, Sunnybrook Research Institute and the University of Toronto, Toronto, Ontario, Canada}
\thanks{Corresponding author: L. Ballerini (email: lucia.ballerini@ed.ac.uk).}}

\markboth{Ballerini \MakeLowercase{\textit{et al.}}: Perivascular Spaces Segmentation in Brain MRI Using Optimal 3D Filtering}{}

\IEEEtitleabstractindextext{%
\begin{abstract}
Perivascular Spaces (PVS) are a recently recognised feature of Small Vessel Disease (SVD), also indicating neuroinflammation, and are an important part of the brain's circulation and glymphatic drainage system. 
Quantitative analysis of PVS on Magnetic Resonance Images (MRI) is important for understanding their relationship with neurological diseases. 
In this work, we propose a segmentation technique based on the 3D Frangi filtering for extraction of PVS from MRI. Based on prior knowledge from neuroradiological ratings of  PVS, we used ordered logit models to optimise Frangi filter parameters in response to the variability in the scanner's parameters and study protocols. We optimized and validated our proposed models on two independent cohorts, a dementia sample (N=20) and patients who previously had mild to moderate stroke (N=48). Results demonstrate the robustness and generalisability of our segmentation method.
Segmentation-based PVS burden estimates correlated with neuroradiological assessments (Spearman's $\rho$ = 0.74, p $<$ 0.001), suggesting the great potential of our proposed method.
\end{abstract}

\begin{IEEEkeywords}
Perivascular Spaces; Ordered Logit Model; Frangi Filter; Brain; MRI
\end{IEEEkeywords}}

\maketitle

\IEEEdisplaynontitleabstractindextext

\section{Introduction}
\label{sec:intro}

\begin{figure}[ht]
\centering
\includegraphics[width=.48\textwidth]{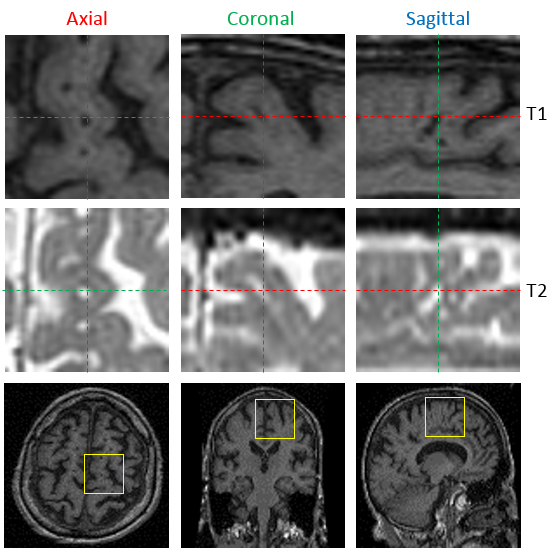}
\caption{\label{fig:PVS_T1_T2} Magnified view of PVS in an axial, coronal and sagittal slice in T1- and T2-weighted MR images. The position of these zooms in corresponding brain scans is highlighted with yellow squares (bottom row).} %
\end{figure}

\IEEEPARstart{P}{erivascular spaces} (PVS), also known as Virchow-Robin spaces, are fluid-filled spaces that follow the typical course of cerebral penetrating vessels. 
PVS have the same Magnetic Resonance Imaging (MRI) contrast characteristics of Cerebrospinal Fluid (CSF), that is they appear hypointense (dark) on T1-weighted (T1) and  hyperintense (bright) on T2-weighted images (T2)~\cite{Wardlaw2013neuroimaging,Ramirez2016imaging}.
They appear as small 3D tubular structures that, depending on the viewing plane, are linear or round, with a diameter generally smaller than 3mm~(see Fig.~\ref{fig:PVS_T1_T2})~\cite{ValdesHernandez2013towards}.

Enlargement of perivascular spaces is associated with other morphological features of Small Vessel Disease (SVD) such as white matter hyperintensities and lacunes~\cite{Doubal2010}; cognitive impairment~\cite{Arba2016} and inflammation~\cite{Potter2015enlarged}.
Most studies use visual rating scales to assess PVS burden~\cite{Potter2015cerebral,Maclullich2004}, but are prone to observer variation, particularly in the Centrum Semiovale (CS), due to the coexistence of PVS with other neuroradiological features of SVD that confound their identification in this region. 

Efforts have been made to computationally assess PVS~\cite{Descombes2004,ValdesHernandez2013towards}.
Recent semi-automatic methods are based on thresholding and require user intervention either for the choice of parameters or for manual editing of the resulting masks~\cite{Ramirez2015visible,Wang2016development}.
One of the most promising approaches proposed for PVS automatic segmentation uses the Frangi filter~\cite{frangi1998multiscale} parameterised through a Random Forest scheme~\cite{criminisi2011decision} that learns discriminative PVS characteristics from manually segmented ground truth on MR images acquired at 7T~\cite{Zong2015visualization,Park2016segmentation}.
However, MRI in clinical research and practice is mostly performed in scanners with field strengths at 1.5T or 3T, and the reference standards available are visual ratings performed by neuroradiologists, which makes the learning-based approach proposed by Park et al.~\cite{Park2016segmentation} limited for practical use.
Moreover, it is difficult to assess enlarged PVS burden at high field 7T MRI since normal PVS and deep medullary veins with similar intensities to PVS confound visualization.
Our current goal is to present a segmentation approach for enlarged PVS that can be used widely in current clinical research studies, to further elucidate their pathological significance and assess their potential role in neurological disorders.

We propose a novel application of ordered logit models, usually used in statistics as a regression model for ordinal dependent variables, as this model provides a good estimate for capturing the sources of influence that explain the ordinal dependent variables (i.e. in this case the PVS visual rating scores) considering the uncertainty (i.e. subjectivity, inter-observer variability) in the measurement of such data~\cite{greene2010modeling}. 
We use this model to estimate the parameters of the Frangi filter~\cite{frangi1998multiscale} to obtain the maximum likelihood of a vessel-like structure to be a PVS in the CS, by also estimating the count of PVS that most likely falls in the class corresponding to the category given by the neuroradiologist in this brain region.

We calibrated different ordered logit model, according to the rating scale available for every dataset. 
We optimized the parameters of the Frangi filter to deal with T1 and T2 modalities, and combined the resulting filtered images. 
Validation was carried out on different cohorts, using images acquired in different sites.

The remainder of the paper is organised as follows: Section~\ref{sec:datasets} introduces the data used in this paper. In Sec.~\ref{sec:methods} we describe the basis of our optimization approach. Section~\ref{sec:implement} discusses some implementation details and the model validation. Following are the validation results which are presented in Section~\ref{sec:results}. Finally the discussion and conclusions are shown in Sec.~\ref{sec:conclusions}.

\section{Materials}
\label{sec:datasets}

Two datasets were used for developing, testing and validating the method:
\begin{enumerate}

\item \textit{Sunnybrook Dementia Study} (SDS): a large ongoing longitudinal study conducted at Sunnybrook Health Science Centre, Toronto, Canada (ClinicalTrials.gov NCT01800214). Patients had an historical profile typical of Alzheimer's disease (AD). Full study details have been published previously~\cite{Ramirez2015visible}.
\item \textit{Mild Stroke Study} (MSS): a study conducted at Centre for Clinical Brain Science, Edinburgh, UK. Patients had clinical features of lacunar or mild cortical stroke. The MRI protocol has been published elsewhere~\cite{ValdesHernandez2015rationale}.
\end{enumerate}

The characteristics of the sequences relevant for PVS assessment are summarized in Table~\ref{tab:protocol}.

\begin{table}[h]
\centering
\caption{\label{tab:protocol} Characteristics of the relevant MRI sequences of (1)~Sunnybrook~Dementia~Study~(SDS)~\cite{Ramirez2015visible}, and (2)~Mild~Stroke~Study~(MSS)~\cite{ValdesHernandez2015rationale}.} 
\begin{tabular}{llll}
\toprule
\multicolumn{2}{l}{Study/parameters} & \textbf{SDS} & \textbf{MSS} \\
\midrule
Matrix & T1 & 256$\times$256$\times$124 &  256$\times$216$\times$256 \\
       & T2/PD$^{*}$ & 256$\times$256$\times$58 & 383$\times$224$\times$28 \\
       & T2cube && 512$\times$512$\times$256 \\
Voxel size & T1 & 0.86$\times$0.86$\times$1.4 & 1.02$\times$0.9$\times$1.02 \\
($mm^3$)  & T2/PD & 0.78$\times$0.78$\times$3 & 0.47$\times$0.47$\times$6 \\
       & T2cube && 0.47$\times$0.47$\times$0.7 \\ 
\midrule
\multicolumn{4}{l}{$^{*}$PD = proton density (interleaved), available in SDS}
\end{tabular}
\end{table}

\section{Methods}
\label{sec:methods}

\begin{figure*}[ht]
\centerline{\includegraphics[width=\textwidth]{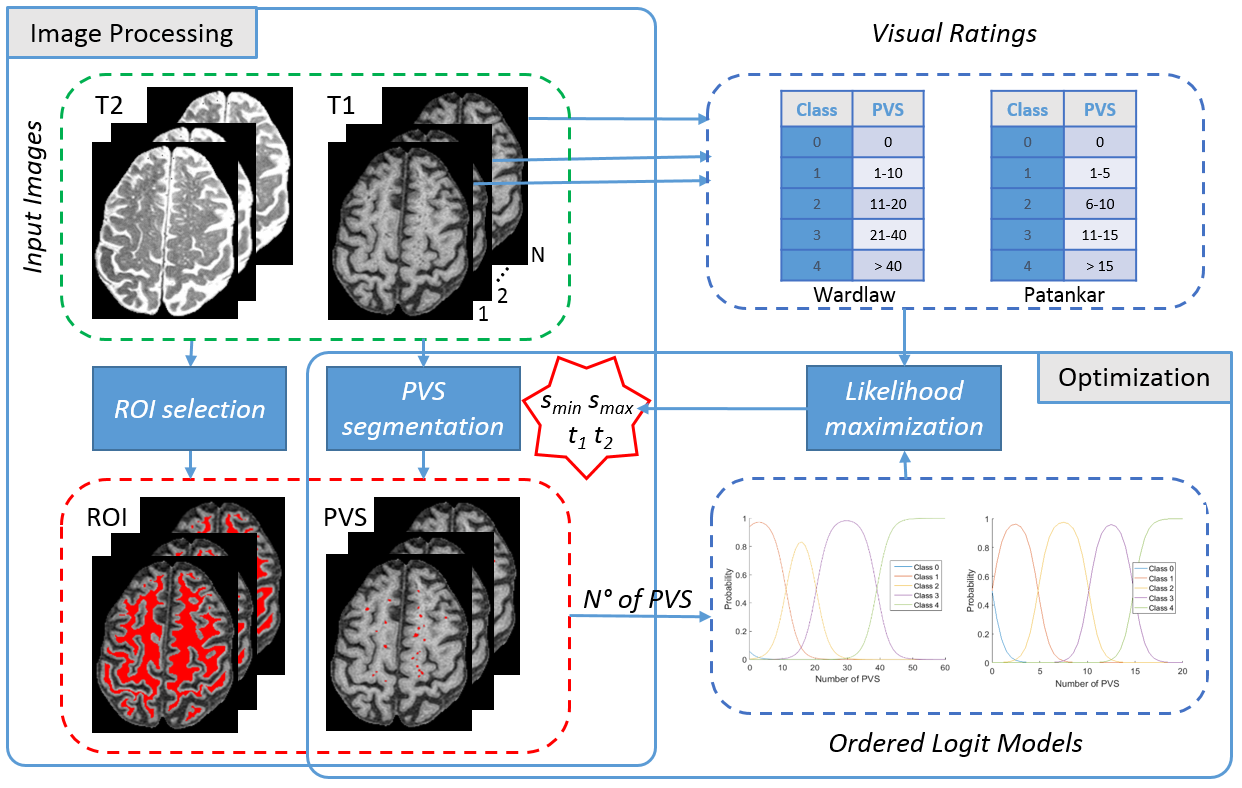}}
\caption{\label{fig:optimization}Flowchart of the proposed optimization approach}
\end{figure*}

Observing the vessel-like structure of PVS, we propose a segmentation technique based on the 3D Frangi filtering~\cite{frangi1998multiscale}, largely used for enhancing blood vessels, for instance in retinal images~\cite{Lupascu2010}. 
Given the absence of an accurate computational "ground truth" (i.e. manual labels of each PVS by experts), we propose a modelling technique to use the available information (i.e. PVS burden assessed using visual rating scales) to optimize the filter parameters. For this scope, an ordered logit model~\cite{greene2010modeling} has been used to simulate the relationship between the number of PVS and the rating categories, taking into account the uncertainty in the measurements.
The flowchart of the proposed optimization process is illustrated in Fig.~\ref{fig:optimization}.

In the remainer of this section we summarize the two visual rating scales used in our work and review the fundamentals of Frangi filtering and ordered logit modelling.

\subsection{Visual Ratings of PVS}
Two established visual rating scales for PVS severity were used in the present work.

The visual rating scale developed by Potter et al.~\cite{Potter2015cerebral} (in the following called Wardlaw scale) required users to rate PVS burden separately on T2-weighted MRI in three major anatomical brain regions: midbrain, basal ganglia and centrum semiovale.
According to the online user guide (http://www.sbirc.ed.ac.uk/documents/epvs-rating-scale-user-guide.pdf), PVS in the latter region should be assessed in the slice and hemisphere with the highest number, and rated as 0 (no PVS), 1 (mild; 1-10 PVS), 2 (moderate; 11-20 PVS), 3 (frequent; 21-40 PVS) or 4 (severe; $>$40 PVS).

On the basis of the measured contrast-to-noise ratios (CNRs), the PVS scores proposed by Patankar et al.~\cite{Patankar2005} were based principally on the appearances seen on inversion recovery images. PVS should be scored in the centrum semiovale as 0 (none), 1 (less than five per side), 2 (more than five on one or both sides).

Two slightly modified versions of these rating methods, as described in~\cite{Ramirez2015visible}, were also used in this work. 
Co-registered MRIs were used for assessment, with T2 for primary identification, T1 for confirmation, and Proton Density (PD) for rejection as required.  To reduce double-counting, a slice increment of 3 was implemented as a standardized rating protocol.  To reduce ceiling effects and account for a greater range of PVS, the Patankar scale was standardized across regions: 0 (none), 1 (one to five), 2 (six to ten), 3  (eleven to fifteen), 4 (sixteen or more).  CS was defined as the White Matter (WM) projections superior to the ventricles, and while both scales normally include sub-insular regions lateral to the lentiform nucleus, we only included the thalamus, lentiform nucleus, caudate nucleus, and internal capsule for evaluation. 

\subsection{Frangi filter}
Frangi~\cite{frangi1998multiscale} analyses the second order derivatives of an image I, defined in the Hessian matrix $H_s(v)$ as: 
\begin{equation}
\label{eq:hessian}
H_s(v)=
\begin{bmatrix} 
I_{xx} & I_{xy} & I_{xz} \\ 
I_{yx} & I_{yy} & I_{yz} \\ 
I_{zx} & I_{zy} & I_{zz} 
\end{bmatrix}
\end{equation}
to describe the "vesselness" $F(v)$ of a voxel $v$ at scale $s$ as: 
\begin{equation}
\label{eq:frangi}
F_s(v)=
\begin{cases} 
0  &  \textit{if} \: \lambda_2 \geq 0 \: \\
& \textit{or} \: \lambda_3 \geq 0, \\ 
(1-e^{-\frac{R_A^2}{2\alpha^2}} ) \cdot e^{-\frac{R_B^2}{2\beta^2}} \cdot (1-e^{-\frac{S^2}{2 c ^2}})  & \textit{otherwise},
\end{cases}
\end{equation}
where $\lambda_1$, $\lambda_2$ and $\lambda_3$ are the ordered eigenvalues $(|\lambda_1| \leq |\lambda_2| \leq |\lambda_3|)$ of the Hessian matrix, $R_A = |\lambda_2|/|\lambda_3|$, $R_B = |\lambda_1|/(|\lambda_2 \lambda_3|)^{1/2}$, $S=(\lambda_1^2+\lambda_2^2+\lambda_3^2)^{1/2}$, and $\alpha$, $\beta$, $c$ are thresholds which control the sensitivity of the filter to the measures $R_A, R_B$ and $S$.

For a bright tubular structure in a 3D image we expect: $|\lambda_1| \leq |\lambda_2|,|\lambda_3|$ and $|\lambda_2| \sim |\lambda_3|$; $|\lambda_1| \sim 0$ and $\lambda_2,\lambda_3 \le 0$.  For a dark structure $\lambda_2,\lambda_3 \ge 0$ and the conditions in Eq.~\ref{eq:frangi} should be reversed.

Given a set of scales $s \in [s_{min},s_{max}]$, the responses are combined as:
\begin{equation}
\label{eq:frangimax}
F(v) = \max_s F_s(v)
\end{equation}
where $s_{min}$ and $s_{max}$ are the minimum and maximum scales at which relevant structures are expected to be found~\cite{frangi1998multiscale}.

\subsection{Ordered Logit Models}

An ordered logit model defines the relationship between an ordinal variable ($y$) which can vary between $0$ and $m (m \in N^+)$, and the vector of independent variables ($x$) by using a latent continuous variable ($y^\star$) defined in an one-dimensional space characterized by threshold points $(\mu_0, \dots , \mu_{m-1})$ as described in equation:
\begin{equation}
y^\star = \beta x + \epsilon, \quad \epsilon \sim G(\mu | \sigma ),  \mu = 0,  \sigma = \pi / \sqrt{3}
\end{equation}
\begin{equation}
\begin{aligned}
&y_{i} = 0 && \textit{if} &&& -\infty < y_{i}^\star \leq \mu_0 \\
&y_{i} = 1 && \textit{if} &&& \mu_0 < y_{i}^\star \leq \mu_1 \\
&\cdots  &&& &&\\
&y_{i} = m && \textit{if} &&& \mu_{m-1} < y_{i}^\star \leq \infty 
\end{aligned}
\end{equation}
where $\beta$ and $\mu_i$ are parameters to be estimated, $\epsilon$ is the error component which has a logistic random distribution with expected value equal to 0 and variance equal to $\pi / \sqrt{3}$, that accounts for the measurement error.
This modelling approach provides a relevant methodology for capturing the sources of influence (independent variables) that explain an ordinal variable (dependent variable) taking into account the measurement uncertainty of such data~\cite{greene2010modeling}.  

Since $y^\star$ is not a deterministic quantity, it is only possible to define the probability to belong to each class:
\begin{eqnarray}
P(y=j|\bar{x}) & = & P(\mu_{j-1} < \bar{y}^\star \leq \mu_j ) = \\
\nonumber & = & L(\mu_j - \beta \bar{x} ) - L(\mu_{j-1} - \beta \bar{x} ), \: j=0 \div m 
\end{eqnarray}
where $L$ is the logistic cumulative distribution function. 

In our work, the ordinal variable ($y$) is the rating class (from 0 to 4) and the independent variables ($x$) is the number of PVS.

\section{Implementation Issues}
\label{sec:implement}

\subsection{Model Calibration}

\begin{figure}[ht]
\centerline{\includegraphics[width=.35\textwidth]{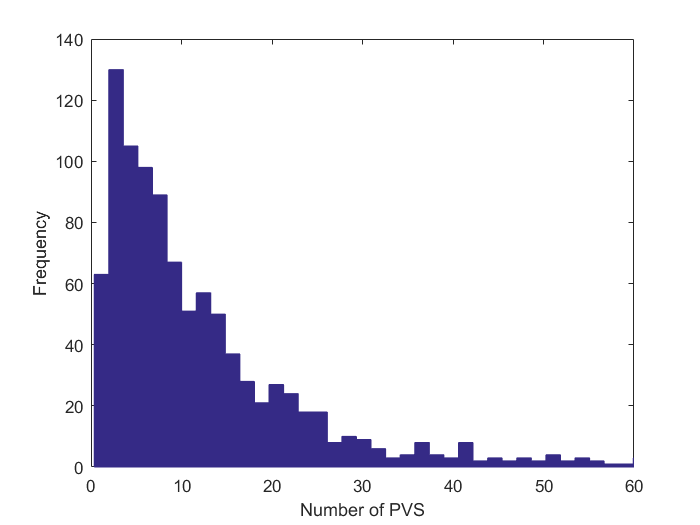}}
\caption{\label{fig:histPVS}PVS distribution of the synthetic dataset generated to calibrate the ordered logit model.}
\end{figure}

The ordered logit model has been calibrated by maximizing a likelihood function based on a synthetic dataset generated in 3 steps.
In the first step 1000 numbers of PVS Count ($PC_i, i=1, \cdots , 1000$) have been generated using a log-normal distribution (see Fig.~\ref{fig:histPVS}), that reflects the observed PVS distribution in known datasets~\cite{Wang2016development}.
In the second step, the uncertainty has been simulated for each $PC_i$ casting a New value of PVS Count ($NPC_i$) using a normal distribution with mean equal to $PC_i$ and standard deviation equal to one. Therefore, the probability that $NPC_i$ is included between $PC_i-3$ and $PC_i+3$ is 0.997. In the third step, a Rating Class ($RC_{ij}$) has been assigned to each generated $NPC_i$.

Assuming $m$ classes, the log-likelihood function can be written as:
\begin{equation}
{LogL}(\mu_i , \beta) = \sum\limits_{i=1}^{1000} \sum\limits_{j=1}^m P( y = j | NPC_j ) RC_{ij}
\end{equation}
where $RC_{ij}$ is equal to one if the $i^{th}$ generated number belong to the $j^{th}$ rating class and it is equal to zero otherwise.

\begin{figure}[ht]%
\centerline{\includegraphics[width=.35\textwidth]{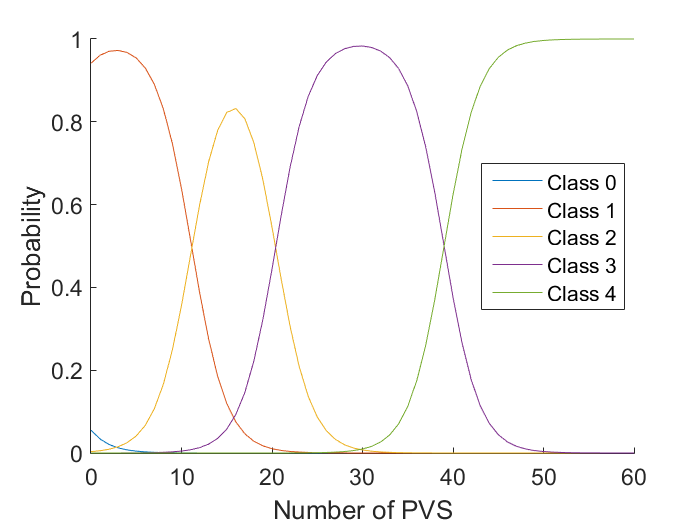}}
\caption{\label{fig:modelWardlaw}Estimated ordered logit model for the Wardlaw rating scale}
\end{figure}

For the Wardlaw scale~\cite{Potter2015cerebral}, a rating class from 0 to 4, being $0 \: (none)$, $1 \: (1-10)$, $2 \: ( 11-20)$, $3 \: (21-40)$, $4 \: (\:>40)$ PVS, has been assigned to each generated number. 
The estimated parameters are $\beta =0.514$, $\mu_0 = -2.840$, $\mu_1 = 5.708$, $\mu_2 = 10.497$, $\mu_3 = 20.040$, and the model is illustrated in Fig.~\ref{fig:modelWardlaw}

\begin{figure}[h]%
\centerline{\includegraphics[width=.35\textwidth]{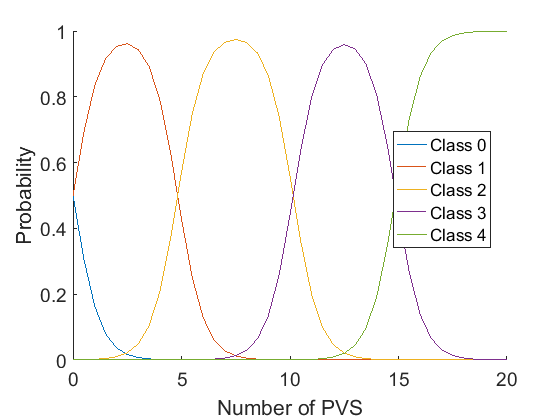}}
\caption{\label{fig:modelPatankar}Estimated ordered logit model for the Patankar rating scale}
\end{figure}

For the modified Patankar scale~\cite{Ramirez2015visible} a rating class from 0 to 4, being $0 \: (none)$, $1 \: (1-5)$, $2 \: (6-10)$, $3 \: (11-15)$, $4 \: (\:>15)$ PVS, has been assigned to each generated number. 
The estimated parameters for the modified Patankar rating scale are $\beta =1.906$, $\mu_0 = 2.269$, $\mu_1 = 9.569 $, $\mu_2 = 18.995$, $\mu_3 = 28.639$, and the model is illustrated in Fig.~\ref{fig:modelPatankar}.

\subsection{Image Preprocessing}
\label{sec:roi}

Images were preprocessed to generate the Region-of-Interest (ROI) masks. 
A fuzzy C-means clustering algorithm was applied to T1 images~\cite{Bezdek1997}.  This is an unsupervised iterative clustering technique that effectively assigns each voxel to one of 4 membership classes: background, Cerebrospinal Fluid (CSF), Gray Matter (GM), and White Matter (WM). After a series of morphological and thresholding operations, the CSF and GM re-labelled voxels were combined to generate the final CSFGM mask which was used for false positive minimization. The CS was identified as the region of WM, superior to the lateral ventricles, present in each of the cerebral hemispheres under the cerebral cortex. In this paper we focused on the CS rather then the basal ganglia (BG), due to the availability of the ROI masks.

PVS masks were also available. These were obtained using Lesion Explorer~\cite{Ramirez2011}, 
which implements 2 false positive minimization strategies: i) in order to reduce errors from minor imaging artifacts and improve differentiation from lacunar infarcts, candidate PVS are required to satisfy acceptance criteria from both T1 and T2, and rejection criteria from PD, and ii) to address potential registration errors and partial volume effects, the cortical GM segmentation from the CSFGM mask was dilated by 1 voxel. This resulted in a relatively conservative estimate of the overall PVS burden and thus, limited its utility as a ground truth for segmentation optimization.

\subsection{Parameter Optimization}
\label{sec:opt}

In order to apply the 3D Frangi filtering, the MRI volumes were first resliced to make $1mm$ isotropic voxels using linear interpolation. 
Then volumes have been filtered according to Eq.~\eqref{eq:frangi} and~\eqref{eq:frangimax} and voxels having F(v) larger than a threshold $t$ were kept.
PVS were identified using 3D connected component analysis as the tubular structures with lengths between 3 and $50mm$~\cite{ValdesHernandez2013towards,Wang2016development}. This provided the initial PVS binary masks.
For each slice we calculated the PVS density as the area of the PVS mask divided by the area of the CS mask (obtained as described in Sec.~\ref{sec:roi}).
We automatically selected the slice in the CS with highest density of PVS.
This slice corresponded to the representative slice having the highest number of PVS selected by the radiologist for assessing the Wardlaw visual ratings~\cite{Potter2015cerebral}.  
The count of PVS in this slice was derived automatically with 2D connected component labelling. 
Similarly, the total number of PVS in the entire CS was obtained with 3D connected component labelling. This count of PVS corresponded to the count performed by the radiologist for the Patankar ratings~\cite{Patankar2005}.   

A log-likelihood function has been defined to optimize the segmentation parameters: Frangi filter scales $s_{min}$, $s_{max}$ and threshold $t$.
In this work, we used the default configuration for the other Frangi filter parameters ($\alpha = 0.5$, $\beta = 0.5$, $c = 500$), as in our previous work,~\cite{BalleriniMIUA2016} we noted that optimizing these parameters produced essentially similar results, at the cost of a much higher computational time. 

Based on the count of PVS ($x_i(s_{min},s_{max},t)$) for each case $i$ we obtained the probabilities of each case $i$ to belong to the five rating classes ($P(y=j|x_i)$, $j=0, \dots , 4$) using the ordered logit model. The PVS visual rating category provided by an expert radiologist was then used to select a probability for each $i$ case ($\bar{P}_i$). The sum of the logarithms of these selected probabilities is the log-likelihood function to maximize:
\begin{equation}
{LogL}(s_{min}, s_{max}, t) = \sum\limits_{i=1}^N log(\bar{P}_i)
\end{equation}
where N is the number of cases.

\subsection{Model Validation}

Segmentation procedures are commonly evaluated by assessing the voxel-wise spatial agreement between two binary masks, one obtained by the automatic method and a manual one. In our case, the manual segmentation of PVS was not available, as it would have been a very tedious and time consuming task to manually annotate these tiny structures in a reasonable size dataset. 
Quantitative comparison with other methods~\cite{Descombes2004,Wang2016development} was unfeasible as they have been applied to MR images having different resolution, acquired using different protocols in different cohorts. 

The performance of the models was therefore validated using Spearman's~$\rho$ (statistical analysis were performed using MATLAB Robust correlation toolbox~\cite{Pernet2013robust}).
Correspondence of PVS total count and volume vs. visual ratings was also assessed to test generalizability.

\section{Experiments and Results}
\label{sec:results}

\subsection{Experiments}

For developing and optimizing the segmentation approach, the imaging datasets of 20 subjects were selected from a sample of the Sunnybrook Dementia Study (SDS)~\cite{Ramirez2015visible}. 

The optimization procedure has been applied to T1-weighted and T2-weighted MRI sequences of the SDS dataset. Frangi filter scales ($s_{min}$ and $s_{max}$) and two thresholds ($t_1$ and $t_2$, one for each modality) have been optimized.  
The range of the parameters that undergo the optimization process has been defined as in Table~\ref{tab:range}.
\begin{table}[h]
\centering
\caption{\label{tab:range}Range of the segmentation parameters to optimize}
\begin{tabular}{lcccccc}
\hline
&  $s_{min}$ & $s_{max}$ & $t_1$& $t_2$ \\
\hline
min value & 0.2 & 2.0 & 0.90 & 0.05 \\
max value & 2.0 & 4.0 & 0.99 & 0.50 \\
\hline
\end{tabular}
\end{table}

PVS were assessed by three raters: two experienced neuroradiologists using the two modified visual rating scales as described in~\cite{Ramirez2015visible}, and a third rater strictly following the guideline of the original Wardlaw~\cite{Potter2015cerebral} and Patankar~\cite{Patankar2005} rating methods. The two ratings (modified Wardlaw and Patankar) of the first raters were close to the conservative estimate of PVS burden obtained as described in sec.~\ref{sec:roi}. The third rater counted all visible PVS in T1 and T2, including the very small ones discarded by the first raters. All raters were blind to each other. Visual ratings are summarized in Table~\ref{tab:ratings}.
\begin{table}[h]
\centering
\caption{\label{tab:ratings}Ratings available for the Sunnybrook Dementia Study (SDS)}
\begin{tabular}{lcccl}
\hline
&  rater 1 & rater 2 & rater 3 & \\
\hline
modified Wardlaw & $\ocircle$ & \checked & & exp 1\\
modified Patankar & $\ocircle$ & \checked & & exp 2\\
\begin{tabular}{@{}c@{}} original Wardlaw \\ original Patankar \end{tabular} & & & 
\begin{tabular}{@{}c@{}} $\ocircle$ \\ \checked \end{tabular} & exp 3 \\
\hline
\end{tabular}
\end{table}

Three sets of experiments were performed as indicated in Table~\ref{tab:ratings}. The symbol $\ocircle$ indicates the scale/rater used for optimization, while the symbol \checked specifies those used for validation. 

\begin{figure}[ht]
\centering
\subfloat[]{\label{fig:Wsmin}\includegraphics[trim={.5cm .5cm .9cm .5cm},clip=false,width=.24\textwidth]{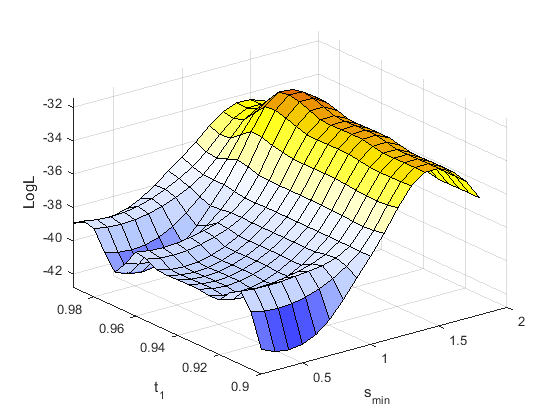}}
\subfloat[]{\label{fig:Wsmax}\includegraphics[trim={.5cm .5cm .9cm .5cm},clip=false,width=.24\textwidth]{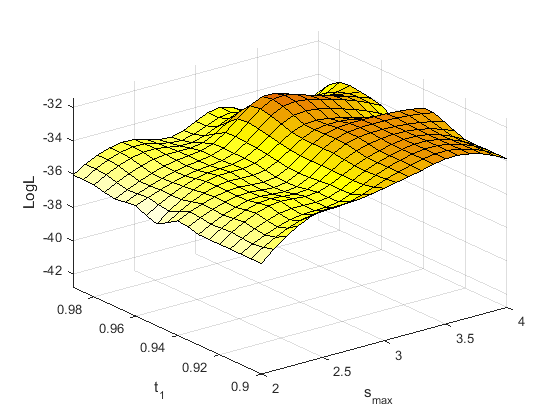}} \\
\subfloat[]{\label{fig:Psmin}\includegraphics[trim={.5cm .5cm .9cm .5cm},clip=false,width=.24\textwidth]{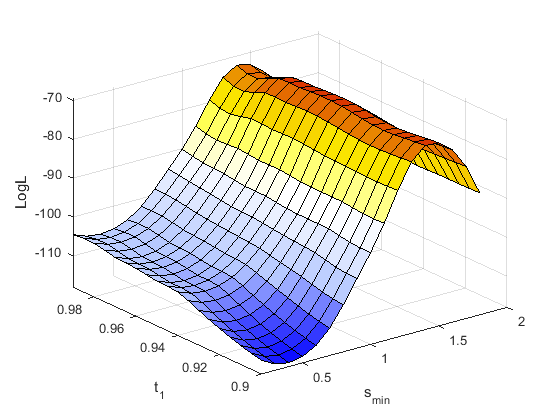}}
\subfloat[]{\label{fig:Psmax}\includegraphics[trim={.5cm .5cm .9cm .5cm},clip=false,width=.24\textwidth]{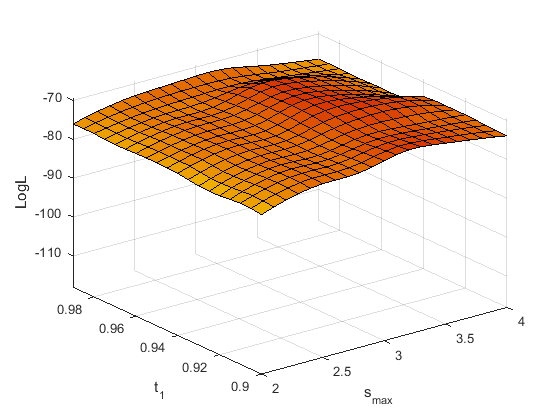}}
\caption{\label{fig:opt}Plots of the log-likelihood function (LogL) for a range of examined $s_{min}$ and $s_{max}$ scales and $t_1$ values using the ordered logit model shown in Fig.~\ref{fig:modelWardlaw} estimated with the modified Wardlaw ratings~(a) and~(b), and using the ordered logit model shown in Fig.~\ref{fig:modelPatankar} estimated using the modified Patankar ratings~(c) and~(d).}
\end{figure}

For illustration Fig.~\ref{fig:Wsmin} and Fig.~\ref{fig:Wsmax} show the surface plots for the parameter optimization using the modified Wardlaw rating scale. Fig.~\ref{fig:Psmin} and Fig.~\ref{fig:Psmax} show the surface plots for the parameter optimization using the modified Patankar rating scale. 
The optimal parameters obtained with the 2 models are very similar ($s_{min}=1.4$, $s_{max}=3.2$, $t_1=0.96$, $t_2=0.35$ for the first model, $s_{min}=1.4$, $s_{max}=3.2$, $t_1=0.95$, $t_2=0.35$ for the second one).

\begin{figure}[ht]
\centering
\subfloat[]{\label{fig:Wt2}\includegraphics[trim={.9cm .5cm .9cm .5cm},clip=false,width=.2\textwidth]{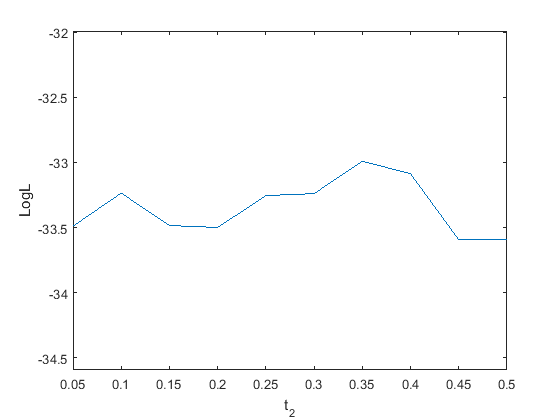}}
\hspace{1cm}
\subfloat[]{\label{fig:Pt2}\includegraphics[trim={.9cm .5cm .9cm .5cm},clip=false,width=.2\textwidth]{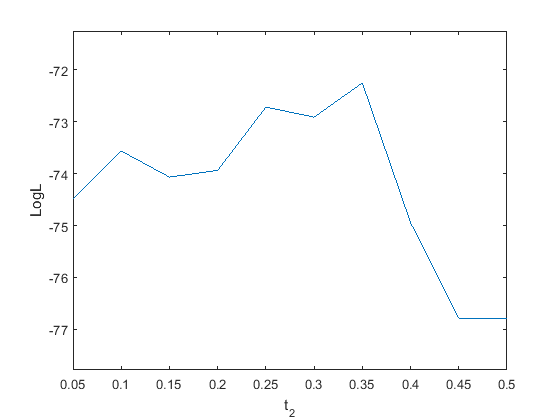}}
\caption{\label{fig:optT2}Plots of the log-likelihood function (LogL) for a range of examined threshold $t_2$ values with the the best combination of $s_{min}$, $s_{max}$ and $t_1$ for the modified Wardlaw~(a) and the modified Patankar~(b) models.}
\end{figure}

From the plots we can observe that the most significant parameter of the Frangi filter is the minimum scale ($s_{min}$). From the figures it is also clear that the Frangi filter was needed. Indeed for any combination of $s_{min}$ and $s_{max}$ the threshold values play a smaller role. 

\begin{figure}[ht]
\centering
\subfloat[]{\label{fig:W3smin}\includegraphics[trim={.5cm .5cm .9cm .5cm},clip=false,width=.24\textwidth]{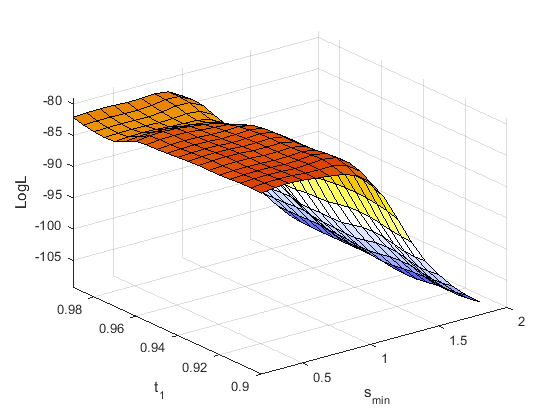}}
\subfloat[]{\label{fig:W3smax}\includegraphics[trim={.5cm .5cm .9cm .5cm},clip=false,width=.24\textwidth]{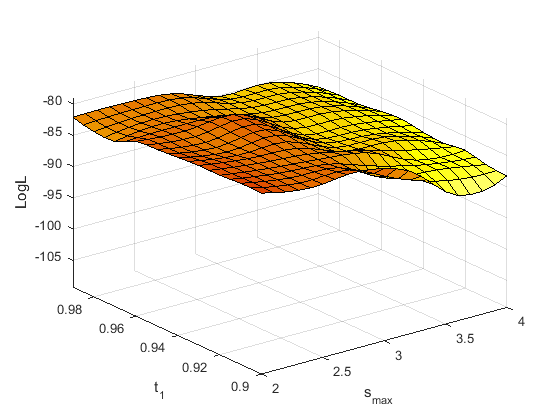}}\\
\subfloat[]{\label{fig:W3t2}\includegraphics[trim={.9cm .5cm .9cm .5cm},clip=false,width=.2\textwidth]{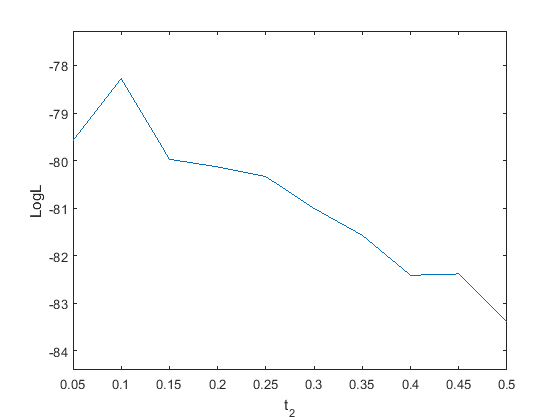}}
\caption{\label{fig:optMVH}Plots of the log-likelihood function (LogL) obtained using the ordered logit model shown in Fig.~\ref{fig:modelWardlaw} estimated with the Wardlaw ratings for a range of examined $s_{min}$~(a) and $s_{max}$~(b) scales and thresholds $t_1$ values, and threshold $t_2$~(c) values with the best combination of $s_{min}$, $s_{max}$ and $t_1$.}
\end{figure}

The optimal parameters obtained with the Wardlaw model using PVS assessed by the third rater where slightly different from the previous ones ($s_{min}=0.2$, $s_{max}=2$, $t_1=0.96$, $t_2=0.1$). The plots of the log-likelihood function are shown in Fig.~\ref{fig:optMVH}.

The trend of plots confirms the validity of the model demonstrating that the model was able to adapt to the rater, and finds the best parameters to segment the PVS accounted by that rater.

\subsection{Qualitative Evaluation}

\begin{figure}[ht]
\centering
\subfloat[]{\includegraphics[trim={14cm 12cm 8cm 12cm},clip=true,width=.157\textwidth]{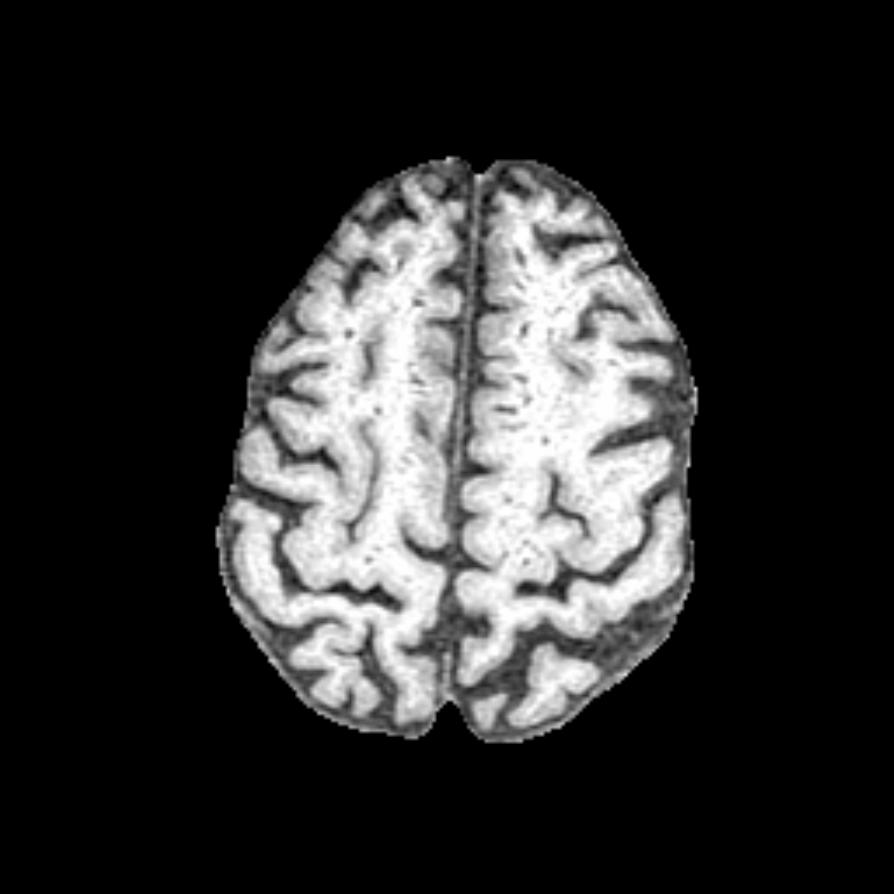}}
\hfill
\subfloat[]{\includegraphics[trim={14cm 12cm 8cm 12cm},clip=true,width=.157\textwidth]{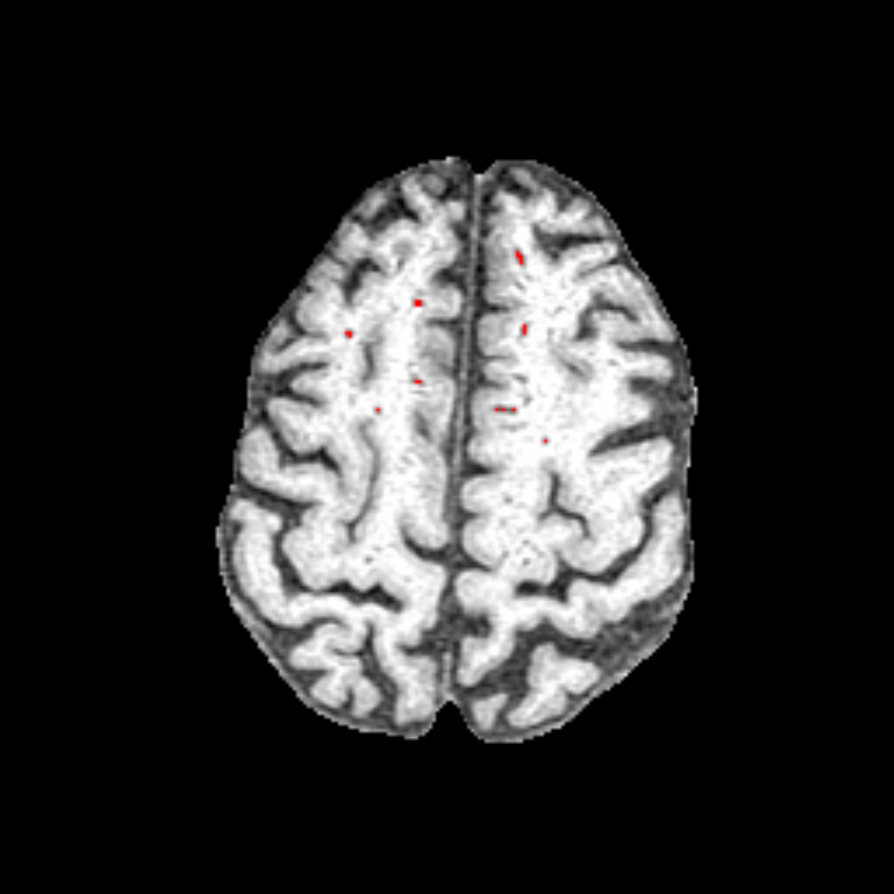}}
\hfill
\subfloat[]{\includegraphics[trim={14cm 12cm 8cm 12cm},clip=true,width=.157\textwidth]{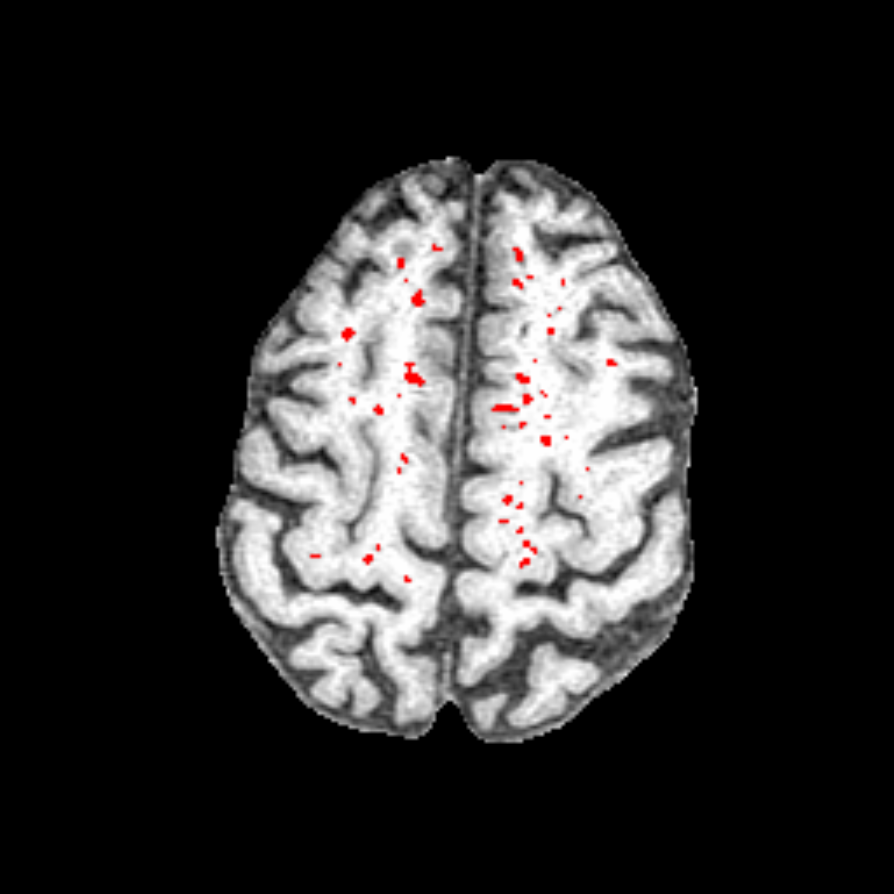}}
\caption{\label{fig:PVScomp}Visual comparison of the PVS segmentation using the conservative threshold based method (b) and the Frangi filtered (c) method.}
\end{figure}

Magnified views of PVS segmentation using a threshold-based method and the proposed method are shown in Fig.~\ref{fig:PVScomp}.

It is clear that the proposed method detected most of the PVS, including the tiny ones, thanks to the enhancement of tubular structure performed by the Frangi filtering using the appropriate scale. The threshold based method missed them, as it was forced to be conservative in order to distinguish PVS from confounding tissue boundaries.

\begin{figure}[ht]
\centering
\includegraphics[trim={7cm 4cm 5cm 5cm},clip=true,width=.157\textwidth]{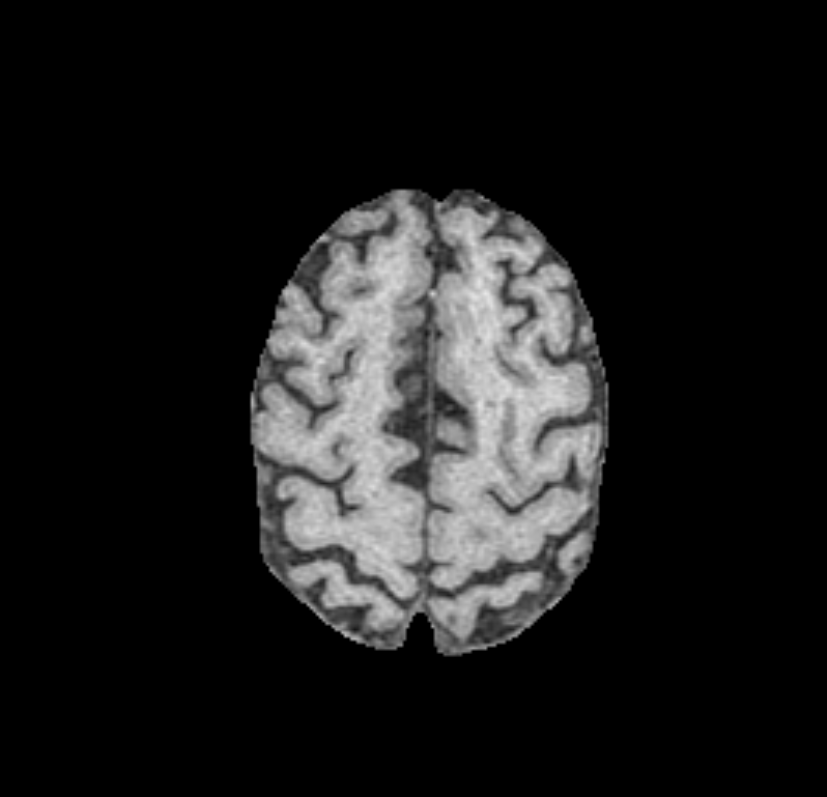}
\includegraphics[trim={7cm 4cm 5cm 5cm},clip=true,width=.157\textwidth]{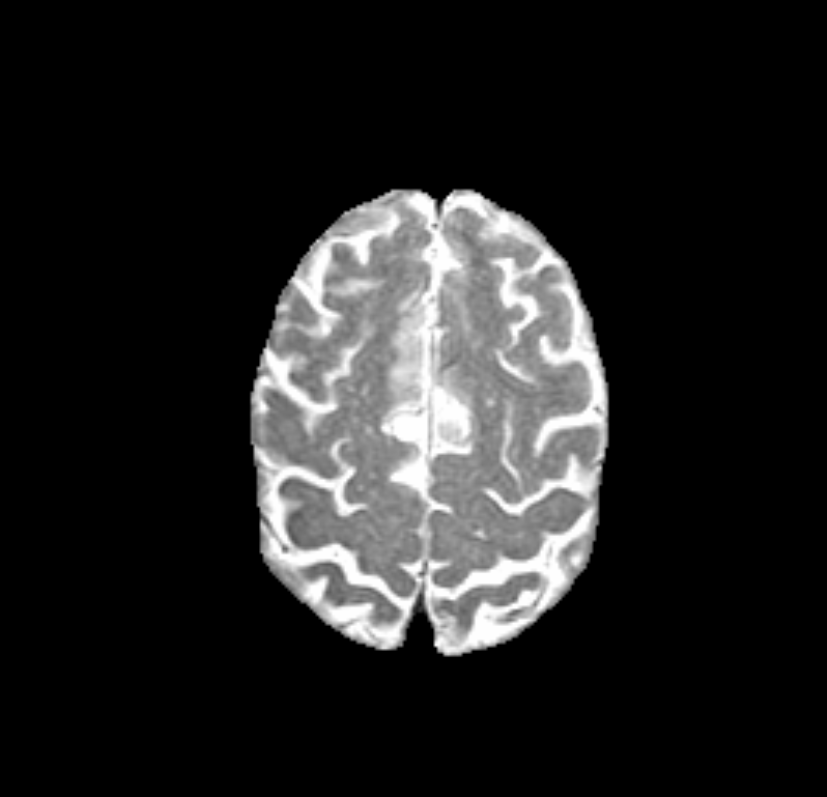} 
\includegraphics[trim={7cm 4cm 5cm 5cm},clip=true,width=.157\textwidth]{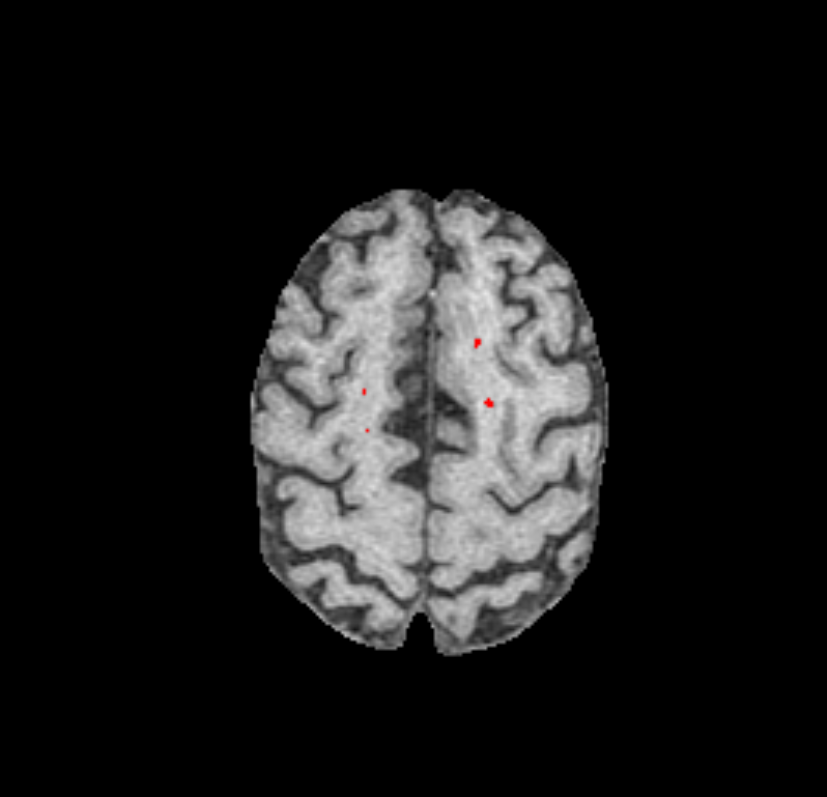} \\
\vspace{.1cm}
\includegraphics[trim={2cm 0cm 6cm 0cm},clip=true,width=.157\textwidth]{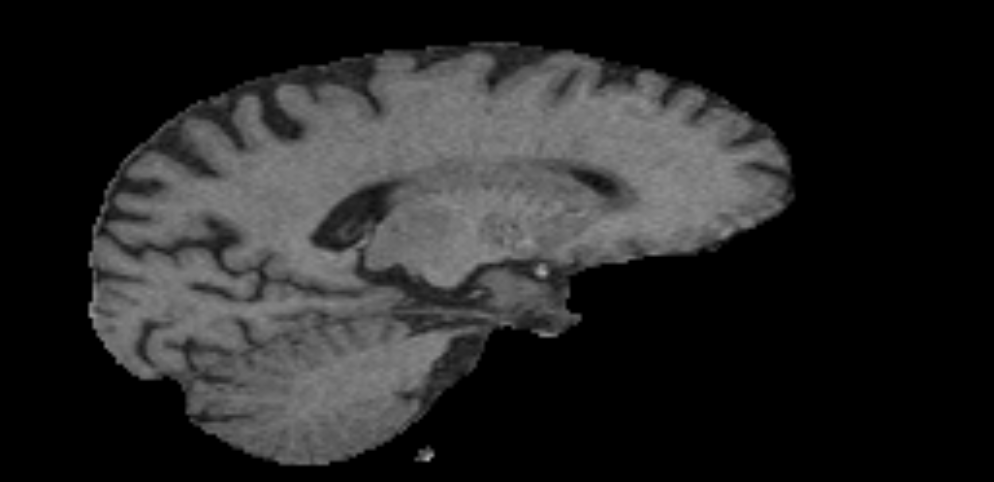}
\includegraphics[trim={2cm 0cm 6cm 0cm},clip=true,width=.157\textwidth]{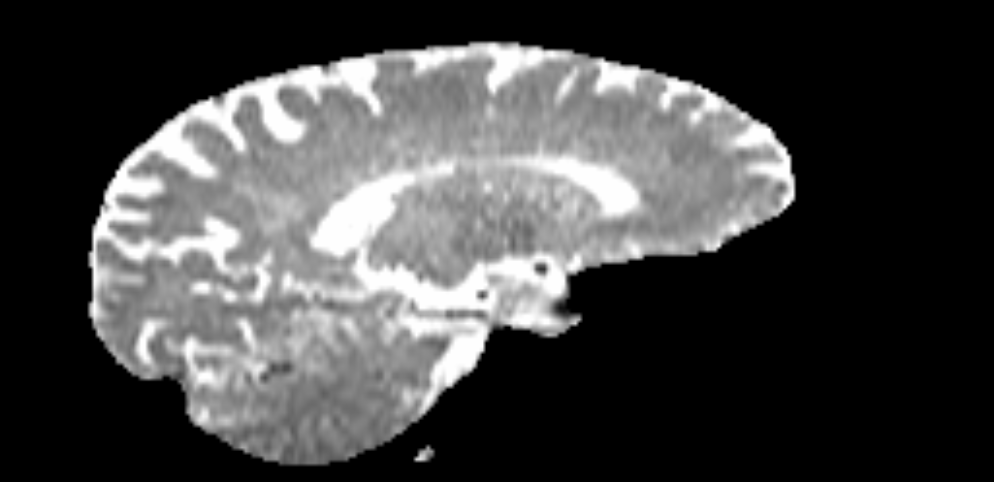} 
\includegraphics[trim={2cm 0cm 6cm 0cm},clip=true,width=.157\textwidth]{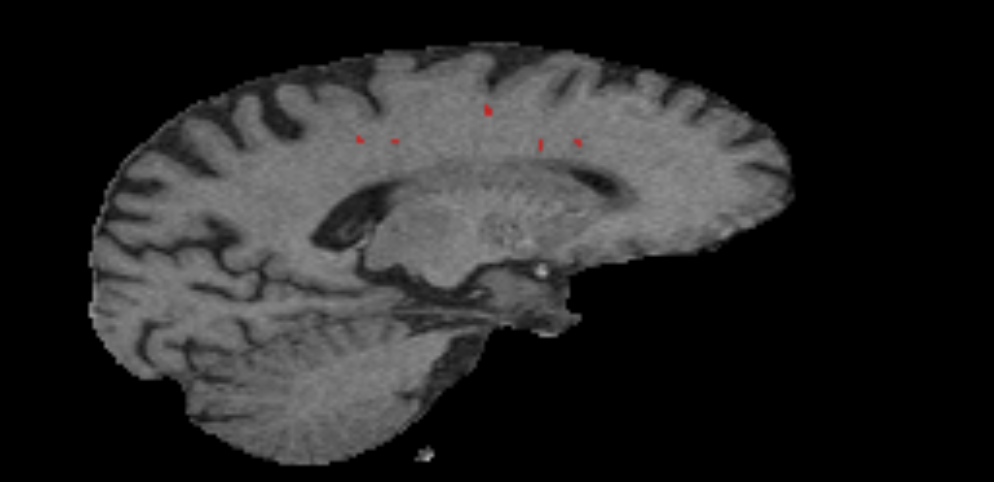}
\caption{\label{fig:PVSsegm1} Examples of the final PVS segmentation a case of SDS dataset having few PVS. Axial (top row) and sagittal (bottom) slice of T1, T2 and PVS overlay (red) on T1. For illustration, T1 are shown in their native space (256$\times$256$\times$124), T2 are shown registered to T1.} 
\end{figure}

\begin{figure}[ht]
\centering
\includegraphics[trim={7cm 3cm 5cm 4cm},clip=true,width=.157\textwidth]{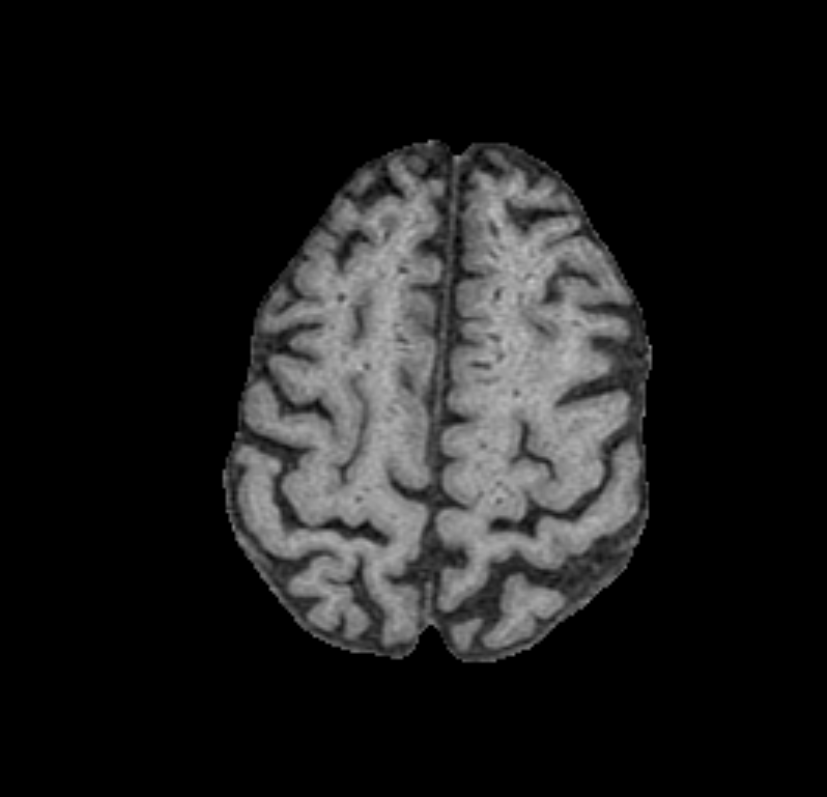}
\includegraphics[trim={7cm 3cm 5cm 4cm},clip=true,width=.157\textwidth]{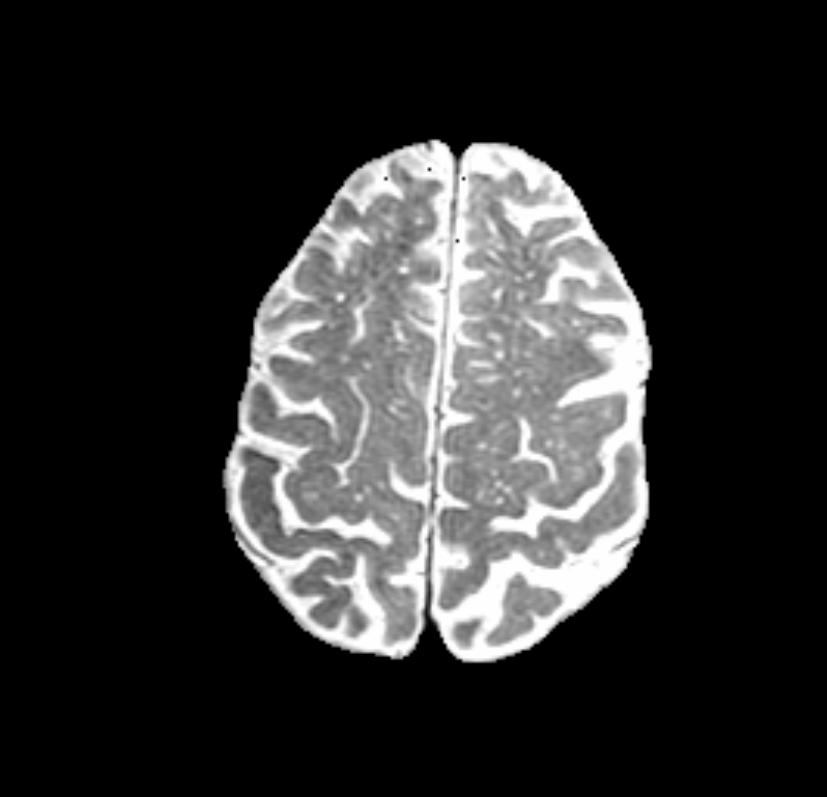}
\includegraphics[trim={7cm 3cm 5cm 4cm},clip=true,width=.157\textwidth]{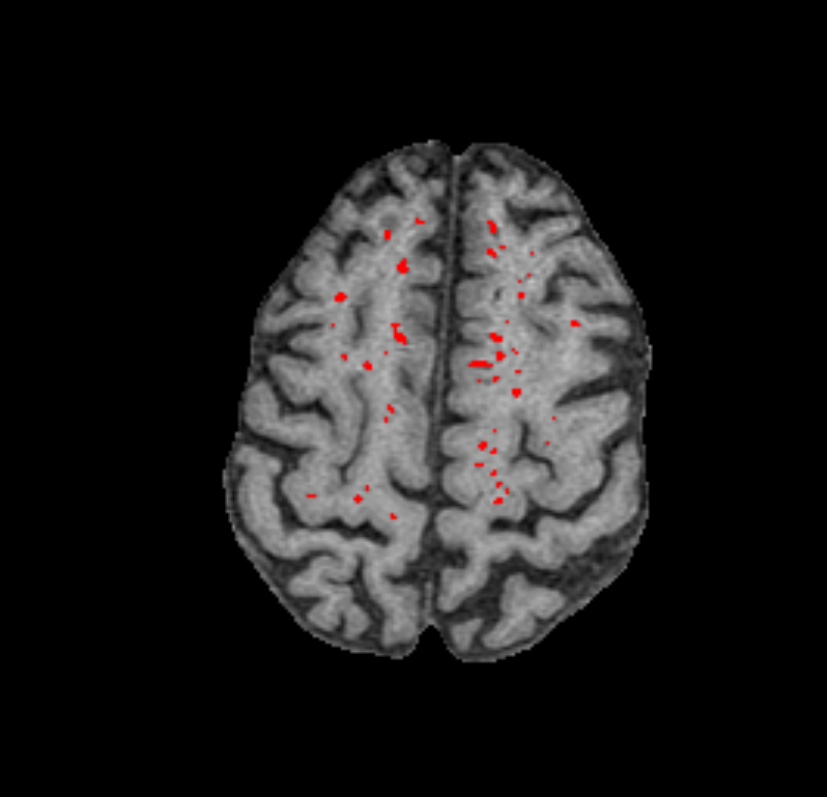}\\
\vspace{.1cm}
\includegraphics[trim={2cm 0cm 4cm 0cm},clip=true,width=.157\textwidth]{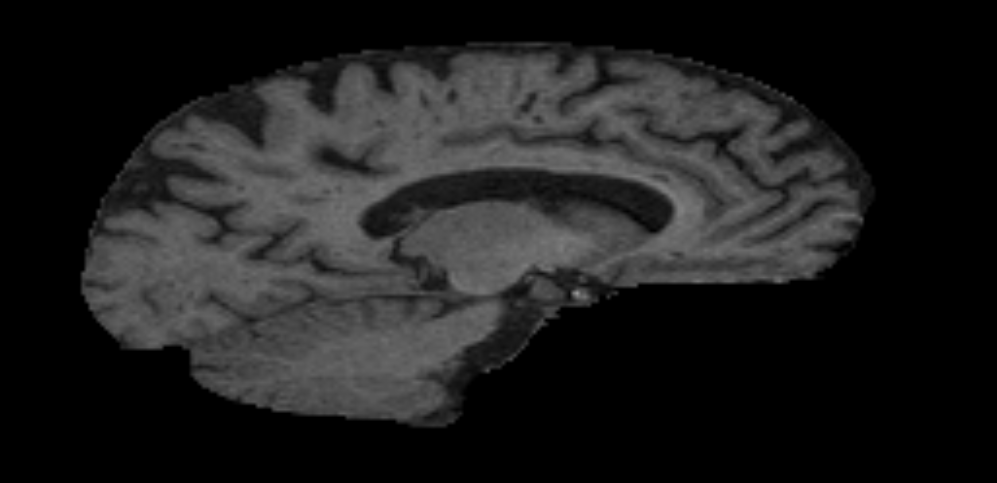}
\includegraphics[trim={2cm 0cm 4cm 0cm},clip=true,width=.157\textwidth]{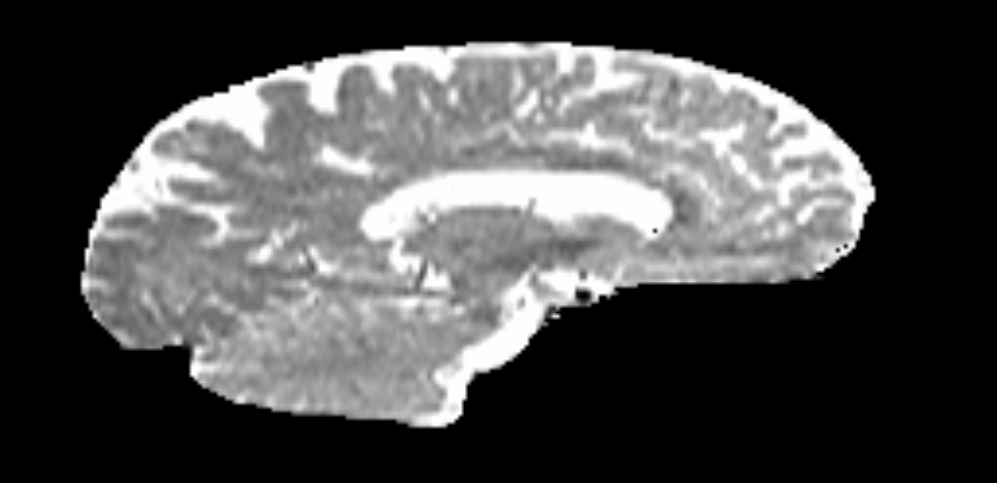}
\includegraphics[trim={2cm 0cm 4cm 0cm},clip=true,width=.157\textwidth]{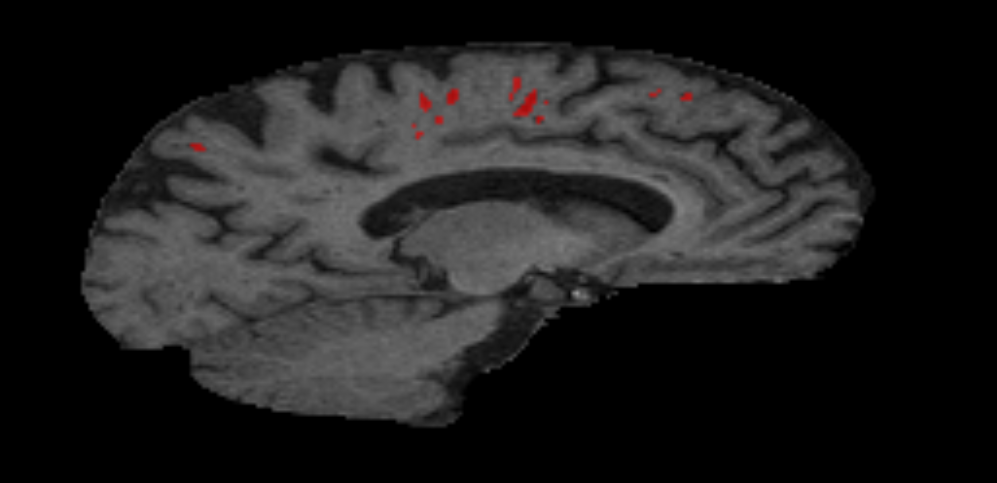}
\caption{\label{fig:PVSsegm2} Examples of the final PVS segmentation for a case of SDS dataset having many PVS. Axial (top row) and sagittal (bottom) slice of T1, T2 and PVS overlay (red) on T1.}
\end{figure}

Examples of segmented PVS for two representative SDS cases having few and many PVS are shown in Fig.~\ref{fig:PVSsegm1} and Fig.~\ref{fig:PVSsegm2}. For each case, we show T1, T2 and the PVS overlay in red.

\begin{figure}[ht]
\vspace{-.35cm}
\centering
\subfloat[]{\includegraphics[trim={3cm 3.3cm 3cm 4cm},clip=true,width=.24\textwidth]{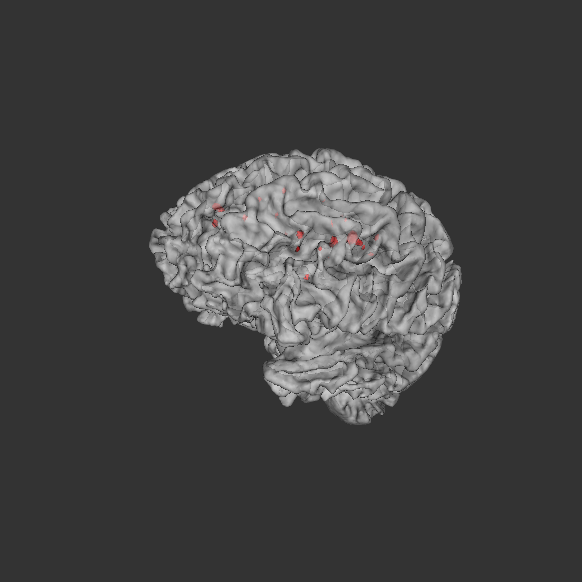}}
\hfill
\subfloat[]{\includegraphics[trim={3cm 4cm 3cm 4cm},clip=true,width=.24\textwidth]{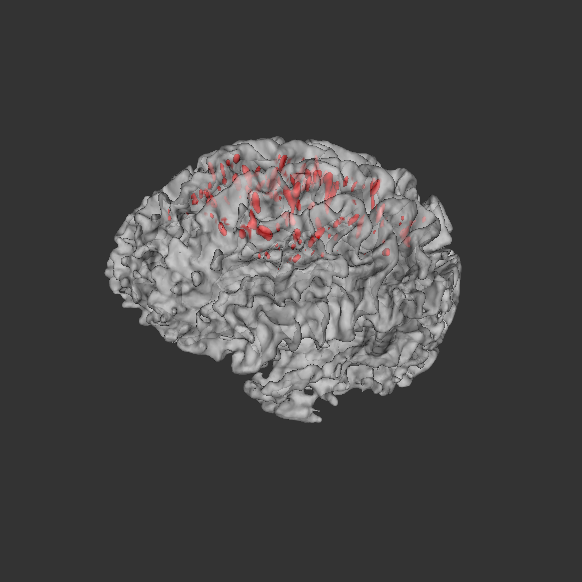}}
\caption{\label{fig:rendering}Volume rendering of segmented PVS (red) for two SDS cases having few (a) and many (b) PVS. PVS volumes overlayed onto a surface render of the brain.}
\end{figure}

Volume rendering of the segmented PVS for two cases having few and many PVS are shown in Fig.~\ref{fig:rendering} for visual qualitative evaluation.

\subsection{Quantitative Evaluation}

When comparing segmentation results with the modified Wardlaw and Patankar visual ratings of the second rater, a high correlation was found for both methods (Spearman's $\rho$ = 0.58, $p=0.006$ and $\rho$ = 0.71, $p=0.0004$ respectively). However, low and no significant correlation was found with total PVS number in or volume CS, suggesting low generalizability. This replicates our previous analysis~\cite{Ramirez2015visible}.

For the segmentation results obtained with the optimal parameters of the model optimized with the original Wardlaw scale a stronger correlation was found (Spearman's $\rho$ = 0.74, $p=0.0002$). 
In addition PVS total count and volume correlates with visual rating scores (Spearman's $\rho$ = 0.67, $p=0.001$ and $\rho$ = 0.53, $p=0.015$, respectively). 

\subsection{Application to alternative acquisitions}

To validate the new PVS segmentation method we applied it to MRI of cases of the Mild Stroke Study (MSS). Visual ratings using the Wardlaw rating~\cite{Potter2015cerebral} were available for all the cases~\cite{Doubal2010}.

Automatic brain, cerebrospinal fluid (CSF) and normal-appearing white matter extraction were performed on T1-weighted MRI using optiBET~\cite{Lutkenhoff14optimized} 
and FSL-FAST~\cite{Zhang2001segmentation} 
respectively. All subcortical structures were segmented, also automatically, using other tools from the FMRIB Software Library (FSL) and an age-relevant template as per the pipeline described elsewhere~\cite{ValdesHernandez2015rationale}.
After identifying the lateral ventricles as the CSF-filled structures with boundaries with the subcortical structures, the CS was identified as the region of normal-appearing white matter, superior to the lateral ventricles, present in each of the cerebral hemispheres under the cerebral cortex. T1-weighted sequence and CS region were linearly registered to the T2-cube images~\cite{Jenkinson2002improved},

The optimization procedure has been applied to T2-cube MRI sequences of 20 patients, and tested on 48 patients  of the same study. 

\begin{figure}[ht]
\centering
\includegraphics[trim={.9cm .5cm .9cm .5cm},clip=false,width=.24\textwidth]{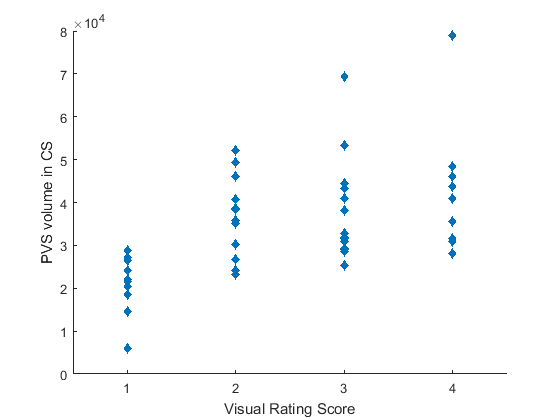}
\includegraphics[trim={.9cm .5cm .9cm .5cm},clip=false,width=.24\textwidth]{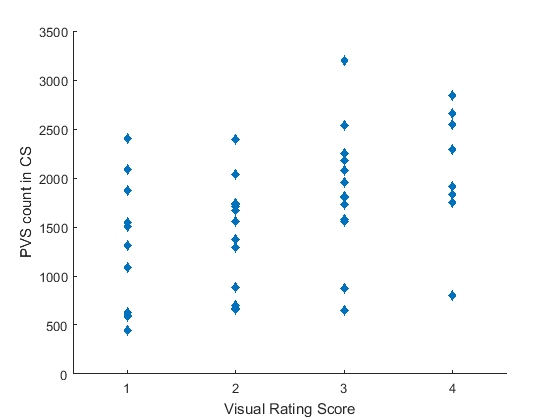}
\caption{\label{fig:association}Associations between PVS computational total volume and count vs. PVS visual rating scores in centrum semiovale (CS) region for the test cases of the MSS dataset}
\end{figure}

PVS total count and volume correlates with visual rating scores (Spearman's $\rho$ = 0.47, $p<0.001$ and $\rho$ = 0.57, $p<0.001$, respectively). 
Scatter plots of these associations are shown in Fig.~\ref{fig:association}. 

The results of this experiment suggest high generalizability of the method. 
CS is more difficult to rate visually than BG, so future application of this method to BG may be more straightforward.

\section{Conclusions}
\label{sec:conclusions}

We presented an automatic method for 3D segmentation of PVS in conventional MRI. 
The 3D Frangi filter enhances and captures the 3D geometrical shape of PVS, thus this method shows promise for identifying and quantifying PVS that run both longitudinally and transversally in the CS, avoiding the double-counting limitations of slice-based methods.
The novelty of this work is the fact that the ordered logit model allows use of the visual ratings for segmentation optimization in absence of alternative computational ground truth. 
The ordered logit model could deal with the measurement uncertainty and the unequal class intervals of the rating scores. 

One limitation of this method is that it relies on the image preprocessing step for the ROI  masks. If the masks provided by this step are not accurate, the method can detect as PVS boundary of grey matter and giri.
Another limitation of this method is that it requires high resolution and quasi isotropic structural MRI. Noisy images have been excluded for this study, otherwise any noise spot of tubular shape can be wrongly segmented as PVS. This can be overcome by a learning method. However, learning methods require GT, and not just visual ratings assessment.

The automatically segmented PVS count and volume agree with visual ratings. 
The method is fully automatic and therefore free from inter- and intra-rater variability. 
However, much more testing is required in a wider range of subjects including those with high burden of other ageing and neuroinflammation features.
The quantitative assessment of PVS volume and count is more suitable for longitudinal studies than visual ratings, that tend to be susceptible to ceiling/flooring effects. 
The accurate segmentation of PVS will allow the analysis of their spatial distribution, orientation and density. Moreover, it will enable the study of the spatial and volumetric relationships of PVS with other markers of SVD, e.g. acute lacunar infarcts, white matter hyperintensities, lacunes, and microbleeds. Additionally, this method shows promise for use in  longitudinal studies where PVS burden can be assessed  in relation to measures of cerebral blood brain barrier permeability, perfusion and cerebrovascular reactivity.
Quantitative measurements will better characterize the severity of PVS in ageing people and their associations with dementia, stroke and vascular diseases. 

This is the first work to propose a multicentre study of PVS segmentation. It shows excellent multi-centre reproducibility.

\section*{Acknowledgment}

L.B. and J.M.W. gratefully acknowledge the financial support from Fondation Leducq Transatlantic Networks of Excellence program (Understanding the role of the perivascular space in cerebral small vessel disease). L.B. gratefully acknowledge the Scottish Imaging Network: A Platform for Scientific Excellence (SINAPSE) for the Postdoctoral and Early Career Exchange award that funded her visit to the Sunnybrook Institute. 

B.J.M., J.R \& S.E.B gratefully acknowledge financial and salary support from the Fondation Leducq, Canadian Institutes of Health Research (\#125740 \& \#13129), Heart \& Stroke Foundation Canadian Partnership for Stroke Recovery, Hurvitz Brain Sciences Research program at Sunnybrook Research Institute and the Linda C. Campbell Foundation. J.R. additionally received partial funding from the Canadian Vascular Network and the Ontario Brain Institute’s Ontario Neurodegenerative Disease Research Initiative. S.E.B. would also like to graciously thank the Sunnybrook Research Institute, Sunnybrook Health Sciences Centre, Dept. of Medicine, and the Brill Chair Neurology, SHSC and Dept. of Medicine, University of Toronto for financial and salary support.

Additionally we would like to thank participants in the studies, radiographers, and staff at Brain Research Imaging Centre, University of Edinburgh and Sunnybrook Research Institute, University of Toronto who contributed to this work.

\bibliographystyle{IEEEtran}
\bibliography{IEEEabrv,PVS}

\end{document}